%% file: main.tex
\documentclass{article}
\usepackage{geometry}
\usepackage{natbib}
\usepackage[utf8]{inputenc} 
\usepackage[T1]{fontenc}    
\usepackage{hyperref}       
\usepackage{url}            
\usepackage{booktabs}       
\usepackage{amsfonts}  
\usepackage{amsmath}
\usepackage{amsthm}

\newtheorem{definition}{Definition}

\usepackage{nicefrac}       
\usepackage{microtype}      
\usepackage{xcolor}         
\usepackage{graphicx}       
\usepackage{subcaption}     
\usepackage{bbm}
\usepackage{tikz}
\usepackage{paralist}
\usepackage{microtype}
\usepackage{bbm}
\usepackage{amssymb}
\usepackage{pdfpages}
\usepackage{fancyhdr}
\usepackage{graphicx}
\usepackage{caption}
\usepackage{subcaption}
\usepackage{multirow}
\usepackage{stmaryrd}
\usepackage[parfill]{parskip}
\usepackage{mathtools}
\usepackage{cases}
\usepackage{comment}
\usepackage{paralist}
\usepackage{breqn}
\usepackage{color}
\usepackage{algorithm, algorithmic}
\usepackage{appendix}
\usepackage{xspace}
\usepackage{enumitem}
\usepackage{gensymb}
\usepackage[english]{babel}
\usepackage{xcolor}
\usepackage{todonotes}
\usepackage{wrapfig}
\usepackage{letltxmacro}
\usepackage{breqn}
\usepackage{caption}
\usepackage{subcaption}

\usetikzlibrary{shapes,decorations,arrows,calc,arrows.meta,fit,positioning,external}
\tikzset{
    -Latex,auto,node distance =1 cm and 1 cm,semithick,
    state/.style ={ellipse, draw, minimum width = 0.7 cm},
    point/.style = {circle, draw, inner sep=0.04cm,fill,node contents={}},
    bidirected/.style={Latex-Latex,dashed},
    el/.style = {inner sep=2pt, align=left, sloped}
}
\definecolor{OI-black}{RGB}{0,0,0}
\definecolor{OI-orange}{RGB}{230,159,0}
\definecolor{OI-lightblue}{RGB}{86,180,233}
\definecolor{OI-green}{RGB}{0,158,115}
\definecolor{OI-yellow}{RGB}{240,228,66}
\definecolor{OI-blue}{RGB}{0,114,178}
\definecolor{OI-vermillion}{RGB}{213,94,0}
\definecolor{OI-purple}{RGB}{204,121,167}

\let\emptyset\varnothing

\LetLtxMacro\oldttfamily\ttfamily
\DeclareRobustCommand{\ttfamily}{\oldttfamily\csname ttsize\endcsname}
\newcommand{\setttsize}[1]{\def\ttsize{#1}}%
\setttsize{\footnotesize}%

\usepackage[symbol]{footmisc}
\input{notation}

\title{Generalizing Off-Policy Evaluation From a Causal Perspective For Sequential Decision-Making}

\author{Sonali Parbhoo$^*$, Shalmali Joshi\footnote{Equal contribution. Authors to prioritise ordering.} \hspace{0.1cm} and Finale Doshi-Velez\\ Harvard University}
\date{September 2021}

\begin{document}

\maketitle

\begin{abstract}
\noindent Assessing the effects of a policy based on observational data from a different policy is a common problem across several high-stake decision-making domains, and several off-policy evaluation (OPE) techniques have been proposed for this purpose. 
However, these methods largely formulate OPE as a problem disassociated from the process used to generate the data (i.e. structural assumptions in the form of a causal graph). 
We argue that explicitly highlighting this association has important implications on our understanding of the fundamental limits of OPE. 
First, this implies that current formulation of OPE corresponds to a narrow set of tasks, i.e. a specific \emph{causal estimand} which is focused on prospective evaluation of policies over populations or sub-populations. 
Second, we demonstrate how this association motivates natural desiderata to consider a more general set of \emph{causal estimands}, particularly extending the role of OPE for \emph{counterfactual} or \emph{retrospective} off-policy evaluation at the level of individual units (e.g. patient-level) of the population. 
Further, a precise description of the causal estimand highlights which OPE estimands are identifiable from observational data under stated generative assumptions. 
For those OPE estimands that are not identifiable from observational data, the causal perspective further highlights where additional experimental data is \emph{necessary} for identification, and thus naturally highlights situations where human expertise can aid identification and estimation. 
Furthermore, many formalisms of OPE overlook the role of \emph{uncertainty} entirely in the estimation process.
We demonstrate how specifically 
characterising the causal estimand highlights the different sources of uncertainty. The role of human expertise then naturally follows through in terms of managing the induced uncertainty. 
We discuss each of these aspects as actionable desiderata for future OPE research at scale and in-line with practical utility.

\end{abstract}

\section{Introduction}

In many real-world applications such as marketing \citep{silver2013concurrent}, healthcare \citep{liao2019off, prasad2019defining, parbhoo2017combining, parbhoo2018improving, gottesman2019combining} and education \citep{mandel2014offline}, it is common to reason about the effects of deploying one policy, based on data collected from a different policy. For instance, in healthcare, given the treatment history of a patient in ICU, we might be concerned about whether patient outcomes would improve if we changed the treatment protocol. Such problems have been posed as off-policy evaluation (OPE) tasks in reinforcement learning, and several methods have been proposed for performing OPE estimation. 

Conventionally, OPE focuses on estimating the population and sub-population-level \emph{prospective} performance of an evaluation policy based on data collected from a different behavior policy~\citep{precup2000,sutton2018reinforcement}. Evaluating prospective performance on (sub)-populations is akin to answering the question \emph{``Will new patients have better outcomes if treated with A instead of B?''}. Yet in practice, we are also interested in questions of the form \emph{``If we had acted differently, would patient X's condition have improved?''}.  
The two key distinctions are that these are inherently require reasoning about \emph{retrospective} performance of an evaluation policy as well as require reasoning on an \emph{individual} level \citep{heckman2001policy, richens2019counterfactual, valeri2016role}.  For prospective reasoning specifically at population and sub-population levels, there is a vast literature in identifying population and sub-population off-policy estimates, including characterizing their statistical properties \citep{dai2020coindice, precup2000eligibility, jiang2016doubly, thomas2016data,levine2020offline}. 
Formalizing the latter (i.e. individual level, retrospective analysis) requires counterfactual reasoning. We argue that there is a need to broaden the traditional purview of OPE to include and enable individual-level reasoning, and formally, counterfactual analysis. We posit that i) this perspective is critical to understanding the limits of current ML-based modeling approaches for OPE, and ii) allows to generalize OPE to answer novel objectives of practical interest. 

We argue that the unifying framework for OPE is to associate the OPE objective to the data-generating mechanism, more specifically, one endowed with an underlying Structural Causal Model (SCM). Using this framework, we can formalize OPE objectives as different \emph{causal estimands} tied to appropriate assumptions over the structure and functional relationships in the associated SCM. Besides this unifying framework, there are several significant implications. First, whether and how well these causal quantities can be estimated from observational data (collected from a different policy) now reduces to figuring out whether the causal estimand is  \emph{observationally identifiable} (i.e., whether we can estimate it from observational data) given our structural assumptions. If identifiable, these necessary assumptions open the possibility of further understanding estimation properties (from a finite number of observational samples). This view also helps formalize OPE if an OPE estimate is non-identifiable from observational data. Specifically, non-identifiability can highlight the need for physical experiments or additional assumptions that could aid identification. Finally, formalizing the causal estimand can help determine the nature of domain expertise required to achieve multiple objectives of identification, estimation, and validating OPE estimates, which can critically improve the practical utility of OPE.
In this position paper, we discuss the need to formalize OPE through a causal lens to advance and generalize OPE to make it a useful tool for practice. 
Our objective is to use the causal perspective to chart a roadmap for generalizing OPE, in particular to expand and address individual level, sub-population level value estimates not just for prospective utility but also for retrospective reasoning in sequential decision-making settings. 
We first cover background focusing on a fundamental sequential decision-making setting of Markov Decision Process and Structural Causal Models in Section \ref{sec:background}. We then show how the conventional view of OPE can be limiting and demonstrate how OPE tasks are formalised as causal estimands in Section \ref{sec:causal_estimand}. In Section \ref{sec:uncertainty}, we discuss how outlining the appropriate causal estimand helps identify different sources of uncertainty, tied directly to i) identifiability of the OPE estimand from observational and/or experimental data, ii) make explicit the modeling assumptions and potential domain expertise required to estimate/ improve statistical properties of intermediate quantities of the OPE estimand. 
In doing so, we subsequently highlight that this characterization naturally formalizes the role of humans in generalizing OPE in Section \ref{sec:humans}. Specifically, we argue that human feedback may be necessary 
in terms of i) providing additional knowledge of the data-generating processes, including highlighting limits of where physical experimentation is possible, ii) explicitly conducting physical experiments to aid identifiability, iii) providing input that can improve statistical properties of OPE estimates in identifiable cases, or iv) providing domain knowledge about potential violations in the knowledge of the data-generating process, such as the amount of potential confounding, thereby enabling validation of OPE estimates. 
When assumptions about the underlying data-generating mechanisms are unclear or need to be relaxed, we argue that this organically generates a desiderata for developing methods that can help overcome various issues in OPE. 
We conclude by providing 
 specific suggestions in the form of a roadmap for using OPE in practice in Section \ref{sec:discussion}.

\section{Background and Notation}
\label{sec:background}
We begin by describing the notion of a Markov Decision Process (MDP) and its role in traditional off-policy evaluation. We subsequently show how augmenting the MDP with an SCM allows us to view different OPE objectives as \emph{causal} estimands tied to data-generating mechanism of the MDP. We use the SCM formulation to formalize various estimands for OPE and discuss implications outlined above in subsequent sections. 
We use capital letters to denote random variables $X$ and small letters to denote their realizations. Domain of a given random variable $X$ is denoted by $\Omega_X$. A collection of random variables is denoted by bold-faced caps letters $\bX$. 

\begin{figure}[t!]
    \centering
    \begin{subfigure}[t]{0.49\textwidth}
    \centering
     \includegraphics{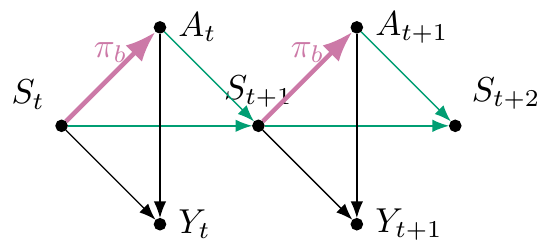}
     \label{fig:mdp_basic_0}
     \caption{}
    \end{subfigure} 
    \begin{subfigure}[t]{0.49\textwidth}
    \centering
     \includegraphics{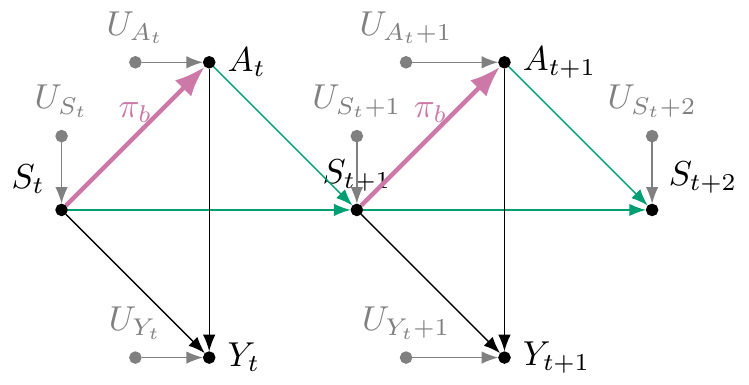}
     \label{fig:mdp_scm_basic}
     \caption{}
    \end{subfigure}
    \caption{MDP and MDP augmented with an SCM. (a): MDP with States $S_t$, Actions $A_t$ and Outcomes/rewards $Y_t$ sampled using behavior policy $\pi_b$. (b): SCM augmented MDP. Gray nodes here are exogenous latents that control the stochasticity of the corresponding node. Pink edges correspond to the behavior policy distribution. Green edges determine the transition dynamics in both figures.}
    \label{fig:mdp_basic}
\end{figure}

{\bf{Markov Decision-Process (MDP) \citep{bellman1957markovian}}.} A canonical Data-Generating Process (DGP) assumed for OPE in reinforcement learning settings is the Markov Decision Process (MDP). MDP (finite or infinite horizon) is defined by the tuple $(\Omega_S, \Omega_A, \cP, \cR, p_0, \gamma)$ where $\Omega_S$ indicates the state-space, $\Omega_A$ indicates the action-space, $P$ denotes the transition dynamics, i.e. $\cP: \Omega_S \times \Omega_A \times \Omega_S \to [0,1]$. The reward is denoted here by $Y \in \Omega_Y$. The reward function governing the MDP is denoted by $\cR: \Omega_S \times \Omega_A \to \Omega_Y$, $p_0$ the initial state distribution and the discount factor $\gamma$. A policy provides a distribution over action given state i.e. functionally, $\pi: \cS \times \cA \to [0,1]$. Let $\cH_{T}^{\pi} \triangleq \{S_0, A_1, Y_1, \cdots, S_{t-1}, A_t, Y_t, \cdots, S_{T-1}, A_T, Y_T \}$ be the trajectories collected under policy $\pi$ and $\cH_{T}^{\pi}$ is the collection of trajectories up to time $T$. Let $G(\cH_T^{\pi}) = \sum_{t=0}^T \gamma^t Y_{t}$. A behavior policy is denoted with the subscript $\pi_b$ and the  evaluation policy is denoted by $\pi_e$. Figure~\ref{fig:mdp_basic}(a) denotes the basic graphical model describing an MDP.





Next we cover background on Structural Causal Models (SCMs) and describe the SCM-augmented MDP that will subsequently be used to generalize OPE. 

\paragraph{Structural Causal Models (SCM) \citep{pearl2009causality}.} A structural causal model $\cM$ describes the causal mechanisms driving a system. It consists of a tuple $\langle \bU, \bV, \bF, \bP\rangle$. Here $\bV$ are the internal or endogenous variables in the causal system and $\bU$ are the set of independent external or exogenous random variables that determine factors of variation in the system; The set of functions $\bF$ govern the causal relationship between a variable $X \in \bV$ and its causal parents $Pa_X$ with the independent exogenous variation due to all exogenous variables $U_X \subseteq \bU$ affecting $X$, given by $X \coloneqq f_{X}(Pa_{X}, U_X)$. The causal dependencies can be summarized in a Directed Acyclic Graph (DAG) $\cG$ with $\bV$ as the nodes and directed edges determined by the SCM. Exogenous nodes are unobserved. Any unobserved nodes between $X$ and $Z$ such that $U_X \cap U_Z \neq \emptyset$, induce dependence or confounding and are usually represented by a  bidirectional edge between $X$ and $Z$. The framework attributes probabilistic (Semi)-Markov assumptions to the joint distribution $\bP$ over the nodes in the graph. This characterizes a probability distribution implying that we can observe samples corresponding to endogenous variables $\bV$ true to the underlying causal graph and mechanism. 

The formulation of structural causal models using the functional relationships and the endowed (Semi)-Markov distribution, allows us to define an intervention in the causal system. An atomic/hard intervention on a node anchors its value to a specific realization (e.g. assigning a treatment to a specific patient), thus removing the causal dependence on the node's parents and any stochasticity due to exogeneous variables. This is denoted as $do(X=x)$ for the node $X$ and replaces the functional assignment $f_{X}(PA_{X}, U_X)$ with $X:= x$. The induced distribution is known as an interventional distribution, denoted by $p(\bV| do(X=x))$. The effect of this is that for all realisations inconsistent with $X=x$, the probability is $0$. This allows for the reasoning: what would be the effect on $\bV \setminus X$ if $X$ is $x$, or practically, what would be the patient outcome if we only gave them an oral medicine.  Note that interventions only affect descendant nodes. In OPE, we are often interested in identifying not just the effect of such a hard intervention, but of an alternative stochastic/deterministic policy, such as giving pills at a slightly lower frequency. These are known as \emph{soft} interventions and result in modifying the functional assignment $f_{X}$ with an alternative governing mechanism $g_{X}$, i.e. $X := g_{X}(Pa_{X}, U_{X})$. A soft intervention is denoted by $\sigma(g_X)$. In this paper, we focus on both soft and hard interventions for a complete exposition. 

\begin{definition}{Soft Interventional Distribution:}\label{def:int_dist}
An intervention $\sigma(g_X)$ on a node $X$ in an SCM $\cM$ consists of replacing the governing mechanism of $X$ given by the function $f_{X}(Pa_X, U_X)$ with a different governing causal mechanism $g_{X}(Pa_X, U_X)$ where $Pa_X$ are the parents of $X$ in a new DAG $\cG^g$. 
\end{definition}

An example of a soft intervention $g$ is 
to evaluate for example, ``what is the effect of treating a patient with pills over shots depending on symptoms?''. Here the mechanism that recommends shots $f$ is replaced by one recommending pills stochastically as a function of patient symptoms. This evaluation is indeed prospective. However, certain questions cannot be answered from such reasoning alone. 
Specifically, causal queries such as ``would the patient's outcome be different if instead they were only provided with medications rather than an invasive procedure?'', are \emph{counterfactual} in nature. 
Counterfactual reasoning can be done for both soft and hard interventions, but requires inferring a model of the exogenous variables $P(\bU | \bV = \bv)$, that informs us of the likely stochasticity under the observational data. This can then be followed by an intervention with $g$ on a causal system with exogenous noise priors $p(\bU)$ replaced by $p(\bU | \bV=\bv)$ (and is usually referred to as the abduction-step). Counterfactuals can therefore be used for \emph{retrospective} reasoning in OPE. 

\begin{definition}{Counterfactual Distribution:}\label{def:count_dist}
Let $\cM_{\bv}$ correspond to the SCM where the exogenous noise model $p(\bU)$ in $\cM$ is replaced by $p(\bU | \bV = \bv)$. Intervening with $g$ on the resulting SCM induces the joint counterfactual distribution $P_{\cM_{\sigma(g)| \bv}}$. The corresponding counterfactual random variable(s) are denoted by $Y_t^{\sigma(g)}(\bU)$ where $\bU$ indicates sampled units that are the target of counterfactual analysis. 
When clear from context we drop the $\sigma(\cdot )$ notation and refer to the variable as $Y_t^{g}(\bU)$. 

\end{definition}

We can augment the MDP (Figure~\ref{fig:mdp_basic}(a)) with an SCM to describe the underlying causal mechanisms governing the sequential decision-making system. The SCM analogue of an MDP can thus be re-stated formally and is denoted in Figure~\ref{fig:mdp_basic}(b). State $S_0$ is a node without parents, and $p_0$ determines the stochasticity of $S_0$. The corresponding exogenous latent $U_{S0}$ is such that $p(U_{S0}) = p_0$, and $S_0 = U_{S0}$ is the nominal functional relationship in the corresponding SCM. The policy $\pi_b$ governs the relationship between actions $A_t$ and its causal parent $S_t$ for all time-steps. That is, $A_t = f_{A_t}(S_t, U_{At})$is governed by $\pi(A_t| S_t)$ where the parameterization of $f$ and dependence on $U_{At}$ is determined by the distribution $\pi_b$. For example, if $A_t$ is normally distributed with mean $S_t$, and variance $\sigma^2_{A_t}$ i.e. $\pi(A_t|S_t) \triangleq \cN(S_t, \sigma^2_{A_t})$, then a potential functional relation in the corresponding SCM is: $A_t = f_{At}(S_t, U_{At}) = S_t + U_{At}$ where $U_{At} \sim \cN(0, \sigma^2_{A_t})$. Similarly the transition dynamics $\cP$ governs the functional relationship $S_{t+1} = f_{S_{t+1}}(S_t, A_t, U_{St+1})$ implicitly. Note that $f$ characterizes the deterministic component and all stochasticity is governed by exogenous $U_{St}$. Similarly for the reward function, i.e. $\cR$ governs the functional relationship $Y_t = f_{Y_t}(S_t, A_t, U_{Yt})$. This perspective leads to an updated interpretation of the canonical OPE as a specific \emph{causal estimand} in an MDP. An example of a soft intervention corresponding to an OPE task is the need to evaluate how an evaluation policy would perform by changing the mechanism governing $A_t$ i.e. $f_{A_t}$, determined by $\pi_{b}$ to an alternative function corresponding to the evaluation policy $\pi_{e}$). 
We formalize this in the next section. 

\section{Formalizing OPE as a causal estimand using SCMs} 
\label{sec:causal_estimand}
\begin{figure}[htbp!]
    \centering
     \includegraphics[scale=0.43]{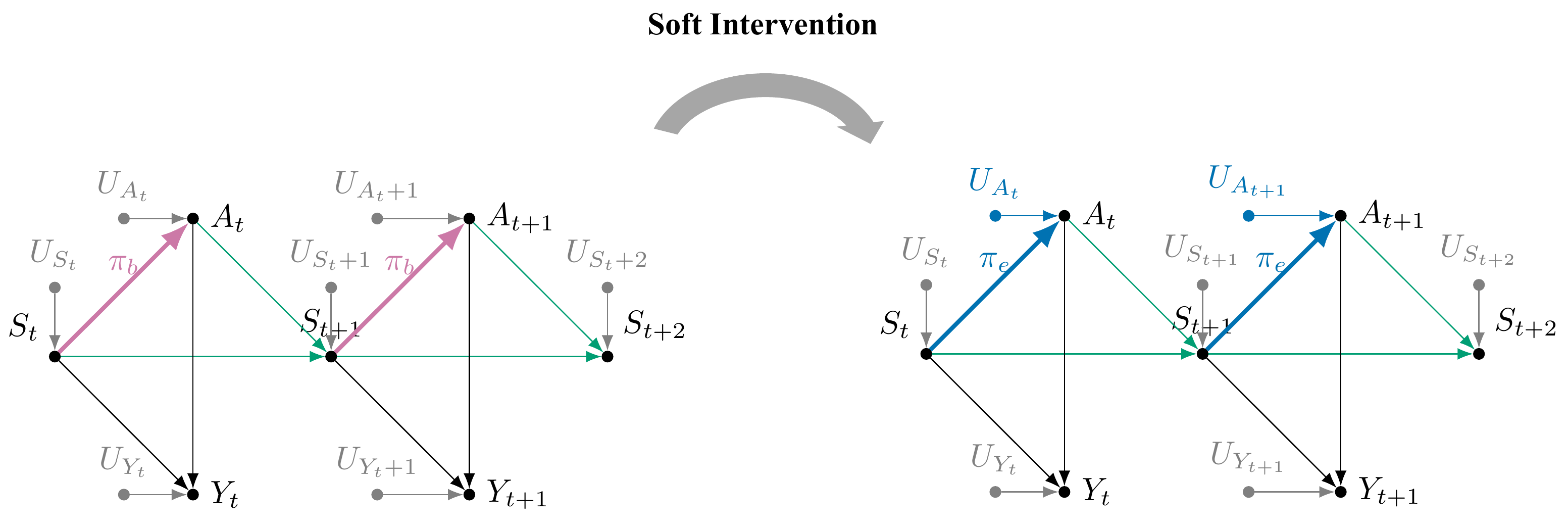}
    \caption{Example of an OPE task as a soft intervention in an MDP where the behavior policy $\pi_b$ is replaced with evaluation policy $\pi_e$.}
    \label{fig:softint}
\end{figure} 




OPE is usually defined in the RL literature as that of estimating the cumulative reward of an evaluation policy $\pi_e$ given observational trajectories from a different behavior policy $\pi_b$. More formally OPE is defined in Definition~\ref{def:ope}.

\begin{figure}[t]
    \centering
    \includegraphics[scale=0.5]{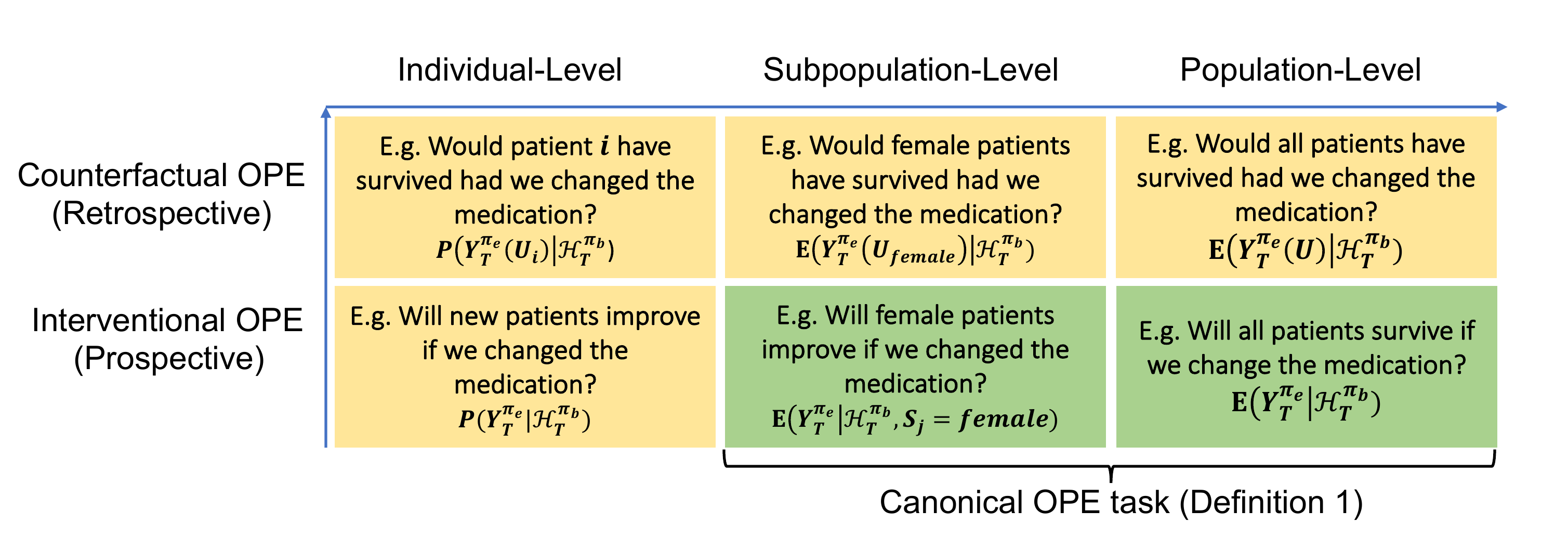}
    \caption{Formalizing OPE through a causal lens along two main axes: i) Estimand type and ii) Counterfactual vs Interventional OPE. Green shades indicate OPE tasks commonly considered in OPE reinforcement learning literature. Yellow shades indicate our proposed generalization by viewing OPE as a causal estimand. Each block contains an example task we might want to accomplish in OPE. Description of the appropriate causal estimand is provided in Section~\ref{sec:causal_estimand}.}
    \label{fig:summary}
\end{figure}

\begin{definition}{Off-Policy Evaluation (OPE) \citep{precup2000}:}\label{def:ope}
Off-policy evaluation estimates the performance $\mathbb{E}_{\pi_e}[G(\cH_{T}^{\pi_e})|\cH_T^{\pi_b}]$ of $\pi_e$ based on a data set of trajectories $\cH_{T}^{\pi_b} \triangleq \{S_0, A_1, Y_1, \cdots, S_t, A_t, Y_t, \cdots, S_{T-1}, A_T, Y_T \}$ each independently generated by different policy(ies) $\pi_b$.
\end{definition}

Definition \ref{def:ope} therefore is focused on i) expected reward from deploying $\pi_e$, and ii) implicitly focused on \emph{prospective} performance of the policy (i.e. on new samples sampled from the data-generating mechanism that deploys $\pi_e$. This is made explicit by augmenting the MDP with an SCM. In the SCM framework, the canonical OPE objective in Definition~\ref{def:ope} corresponds to a \emph{soft intervention} of the form in Figure~\ref{fig:softint} (shown here for the standard MDP case). This canonical definition corresponds to the causal estimand $\mathbb{E}[G(\cH_{T}^{\pi_e})| \cH_{T}^{\pi_b}]$ or  when interested in the final outcome, just $\mathbb{E}[Y_T^{\pi_e}| \cH_T^{\pi_b}]$. 

{\bf{Prospective OPE and interventional estimation.}} To make decisions about the utility of a policy at an individual level, we are not only interested in expected outcomes but on the probability distribution of the outcome of interest. For example, we may want to answer: \emph{``Will new patients improve if we changed the medication?''.} The corresponding causal estimand is $p(Y_{T}^{\pi_{e}}| \cH_{t}^{\pi_b})$ deviating from the canonical OPE estimand. Subgroup-level OPE estimation further demands conditional expectations based on specific attributes of interest: \emph{``Will female patients improve if we changed the medication?''} and corresponds to the causal estimand $\mathbb{E}[Y_T^{\pi_e}| \cH_T^{\pi_b}, S_{:,j} = female]$ where $j$ indexes a specific attribute in the state representation vector (and is usually invariant over time hence not indexed by $t$. A more significant distinction which is excluded from the purview of OPE is whether one is interested in retrospective OPE i.e. answering whether for example \emph{``Would this patient have survived had we changed the medication?''} This distinction is made clear in the SCM augmented MDP and the intuition of what it means to conduct such counterfactual reasoning. 

\begin{figure}[htbp!]
    \centering
    \includegraphics[scale=0.4]{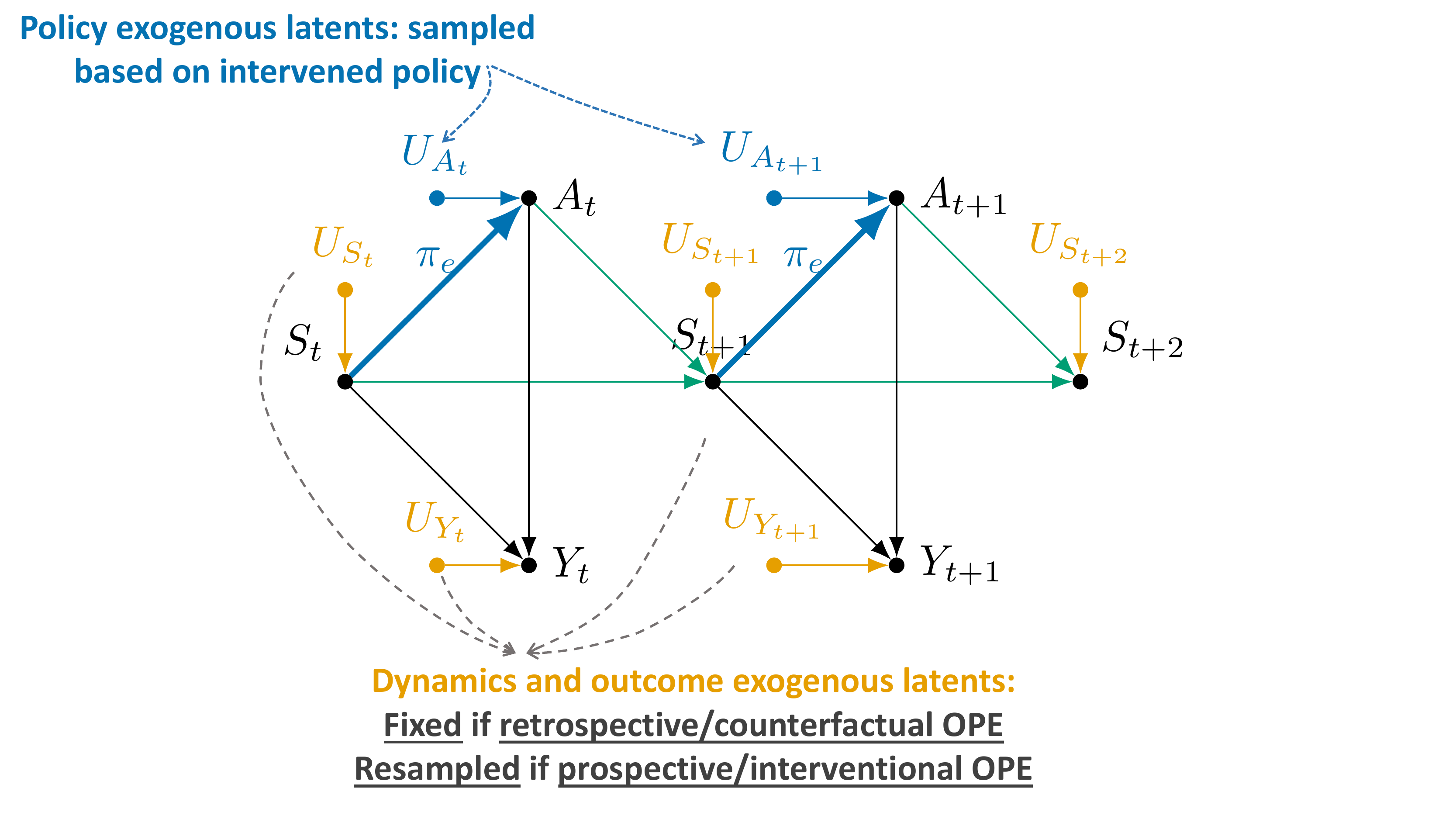}
    \caption{Distinguishing  Interventional (prospective), and Counterfactual (retrospective) OPE,  using distinct causal estimands. This figure depicts the distinction in the inference procedure (post-soft-intervention using the evaluation policy) for each case. For interventional OPE or prospective off-policy tasks, \emph{new} units (samples from the MDP) are the target of inference. The appropriate estimand can be individual-level, sub-population-level or population-level (and can be appropriately conditioned on post re-sampling during inference). For counterfactual OPE or retrospective off-policy tasks, appropriate unit of inference (specific individual/unit) or a sub-group of individuals (sub-population units) are held \emph{fixed} for inference.}
    \label{fig:comparison}
\end{figure}

First, any OPE task, requires a soft intervention for either retrospective or prospective analysis i.e. a using policy $\pi_e$ instead of $\pi_b$, thus changing the mapping $S_t \to A_t$ (denoted in Figure~\ref{fig:softint}). To sample actions, we this sample exogenous variables $U_{At}$'s according to the new policy mapping in either the retrospective or prospective OPE task. The main distinction is in the choice of units leveraged for inference over patient transitions and outcomes. Since the goal of interventional or prospective OPE tasks is to  evaluate utility of a policy on \emph{new} units or subpopulations post-soft-intervention,  this inference is conducted on units re-sampled from the prior indicating prospective units. The OPE task can be individual-level, sub-population-level or population-level and can be estimated with appropriate conditioning post re-sampling. We make these distinctions clear in  Figure~\ref{fig:comparison}. A unit corresponds to a sample associated with the MDP-SCM.  The queries associated with this task are summarized in the bottom row of Figure~\ref{fig:summary}. For individual level prospective OPE, we focus on the $P(Y_T^{\pi_e} \vert \cH_T^{\pi_b})$, i.e. the prospective outcome after intervening with $\pi_e$ using data collected from $\pi_b$. For sub-population-level estimands, we appropriately conditioned to identify units of interest (e.g. females) in Figure~\ref{fig:summary}. 

{\bf{Retrospective OPE and counterfactual reasoning.}} On the other hand, for retrospective OPE, requiring counterfactual reasoning, the target unit(s) are fixed, identified by the task e.g. a fixed patient for individual-level reasoning, all females we have observed so far for sub-population counterfactual OPE and all units observed so far for population-level counterfactual OPE. To reason about the fixed patient then, the exogenous latents corresponding to the unit are held fixed (for unit $i$, we restrict to exogenous $\bU_i$. Similarly for sub-population-level estimation (e.g. females, we restrict to female units denoted by $\bU_{female}$ over which we take expectations post-intervention.  The top row in Figure~\ref{fig:summary} highlight the corresponding causal estimand ($p(Y_T^{\pi_e}(U_i) \vert \cH_{T}^{\pi_b})$ for individual-level counterfactual OPE and $\mathbb{E}_{\bU_{female}}[Y_T^{\pi_e}(\bU_{female})| \cH_T^{\pi_b}]$ for sub-population-level counterfactual OPE). Figure~\ref{fig:comparison} highlights how units are appropriately fixed, while the soft-intervention using  $\pi_e$ implies the actions will be sampled according to the intervened policy.

In the following we focus on two tasks (one retrospective OPE and one prospective OPE) to demonstrate that effectively addressing all OPE tasks require assumptions on the data-generating process, outlining conditions that result in identifiability from observational data, which also further determine the intermediate quantities that may be required for estimation, and finally requirements for validation of OPE.

\paragraph{Example 1: Will all patients survive if we change the medication?} 
This OPE task is corresponds to a population-level prospective OPE estimand (green box) in Figure~\ref{fig:summary}.  
As described in previous sections, evaluating the (changed medication policy) based on data collected from (an old medication policy) corresponds to a soft intervention that modifies the policy mechanism. As a prospective task, inference requires sampling exogenous mechanisms from the prior that would allow to sample new units over which we take an expectation to get a population level estimate.  
Nonetheless, apriori, it is unclear whether observational data $\cH_{T}^{\pi_b}$ is sufficient to estimate the corresponding causal estimand, $\mathbb{E}[\sum_{t=0}^T\gamma^t Y_{t}^{\pi_e}| \cH_{T}^{\pi_b}]$ (generally, but also in the specific model shown in Figure~\ref{fig:softint}). 

We start with assuming that the data is generated from an MDP (Figure~\ref{fig:softint}). We explain specific challenges using the example of Importance sampling (IS), which is often used to get an unbiased estimate of this estimand, when data is collected using policy $\pi_b$. Below we demonstrate the IS-OPE estimand~\citep{precup2000eligibility} and highlight assumptions made along the way that relate to and can be implied from assumptions on the data-generating mechanism. 
\begin{align}
     \mathbb{E}_{\bS, \bY}[\sum_{t=0}^T \gamma^t Y_t{^{\color{OI-blue}\pi_e}} | \cH_T{^{\color{OI-purple}\pi_b}}]  
     &= \mathbb{E}_{\bS, \bY, A_t \sim {\color{OI-blue}\pi_e}}[\sum_{t=0}^T \gamma^t Y_t^{A_t} | \cH_T{^{\color{OI-purple}\pi_b}}]  \\
     &= \int  \sum_{t=0}^T \gamma^t Y_t^{A_t}\prod_{t=1}^T p(Y_t| A_t, S_t){\color{OI-blue}{\pi_{e}(A_t|S_t)}} {\color{OI-green}{p(S_{t+1}|A_t, S_{t})}} p(S_0) \label{eq:dgm}\\
     &= \int  \sum_{t=0}^T \gamma^t Y_t\prod_{t=1}^T p(Y_t| A_t, S_t){\color{OI-blue}{\pi_{e}(A_t|S_t)}} {\color{OI-green}{p(S_{t+1}|A_t, S_{t})}} \label{eq:ignorability} p(S_0) \\
     &= \int \sum_{t=0}^T \gamma^t Y_t \prod_{t=1}^T p(Y_t| A_t, S_t){\color{OI-blue}{\pi_{e}(A_t|S_t)}} {\color{OI-purple}{\frac{\pi_{b}(A_t|S_t)}{\pi_{b}(A_t|S_t)}}} {\color{OI-green}{p(S_{t+1}|A_t, S_{t})}} p(S_0)\\
     &=\int \sum_{t=0}^T \gamma^t Y_t \prod_{t=1}^T p(Y_t| A_t, S_t){\color{OI-purple}{\pi_{b}(A_t|S_t)}} \frac{{\color{OI-blue}{\pi_{e}(A_t|S_t)}}}{{\color{OI-purple}{\pi_{b}(A_t|S_t)}}} {\color{OI-green}{p(S_{t+1}|A_t, S_{t})}} p(S_0)\\
      &= \int \underbrace{\sum_{t=0}^T \gamma^t Y_t \prod_{t=1}^T p(Y_t| A_t, S_t){\color{OI-purple}{\pi_{b}(A_t|S_t)}}  {\color{OI-green}{p(S_{t+1}|A_t, S_{t})}}p(S_0)}_{\text{Trajectories under behavior policy}} \underbrace{\frac{{\color{OI-blue}{\pi_{e}(A_t|S_t)}}}{{\color{OI-purple}{\pi_{b}(A_t|S_t)}}}}_{\text{IS Re-weighting factor}} \label{eq:overlap}
\end{align}

Equation~\ref{eq:overlap} shows importance re-weighting used on trajectories sampled from $\pi_b$ in order to obtain OPE estimates corresponding to our target population-level estimand. From the derivation, we re-iterate and highlight assumptions~\cite{precup2000}. First, Equation~\ref{eq:dgm} implies a factorization of the data-generating process which corresponds to the MDP assumption in Figure~\ref{fig:softint}. Equation~\ref{eq:ignorability}, replaces interventional random variables with observed counterparts due to sequential ignorability, which follows from the structural assumption of the MDP and the backdoor-criterion~\citep{richardson2013single}: $Y_{t}^{a_t}, Y_{t}^{a'_t}, \cdots  \perp A_t \vert \cH_{t}$. Finally, Equation~\ref{eq:overlap}, is estimable due to assumptions of  overlap: that the support of $\pi_e(A_t| S_t) > 0$ wherever $\pi_b(A_t| S_t)> 0$. 
More specifically, when  we assume ignorability conditioning to $S_{t}$, that is, $Y_{t}^{a_t}, Y_{t}^{a'_t}, \cdots  \perp A_t \vert S_{t}$, we assume that the underlying DGP corresponds to the MDP as in Figure~\ref{fig:softint} (left). When these assumptions hold, $\mathbb{E}[G(\cH^{\pi_e})|do(\pi_e), \cH_{T}^{\pi_b}]$ is estimable as we show above and IS provides an unbiased estimate of our OPE estimand.


Equation \ref{eq:overlap} therefore demonstrates how the causal estimand can be converted into a quantity estimable from observational data, albeit under the assumptions we highlighted. More importantly, we demonstrate that IS estimates of OPE are making implicit assumptions about observing all sources of confounding, overlap, and focusing on an expected measure of the value. When attempting prospective OPE evaluation for sub-groups, we further need to appropriately condition on (i.e. take expectations over sub-group identifying properties such as females). When interested in \emph{individual-level} prospective OPE estimation, we may be interested in estimating not just the expected cumulative outcome but the complete distribution of the long-term come, thus changing the estimand. The issue of identifiability is implicit in the assumptions of sequential ignorability. When any parts of the state-space may be unobserved, this assumption may not be satisfied thus rendering IS-based prospective OPE estimates biased (see Figure~\ref{fig:pomdp} for an example and Section~\ref{sec:learning} on handling non-identifiability).  We now turn a counterfactual OPE estimand to highlight how assumptions tied to the structure and functional relationships thereof can make identifiability (as well as  estimation and validation) challenges transparent. 


\begin{align}\label{eq:ett1}
 \begin{split}
\mathbb{E}[Y_t^{A_t=a_t'}] &= \sum_{a \in \{a_t,a_t'\}} \mathbb{E}[Y_t^{A_t=a_t'}| A_t = a] p(a) \\
    \end{split}
\end{align}
\paragraph{Example 2: Would all patients have survived had we changed the medication?} This OPE task, as pointed in out in Figure~\ref{fig:summary} (yellow box on top right), is a retrospective OPE  task at the population-level.  
The most common example of this is the expected causal Effect of the Treatment on the Treated or ETT~\citep{heckman1992randomization}, denoted by $\mathbb{E}[Y_t^{A_t=a_t'}|A_t=a_t]$ (where $a_t$ is the original medication and $a_t'$ is the changed medication. Specifically, this evaluates the expected outcome had the patients been treated with $a_t'$ instead of $a_t$. Note that this is an example of a hard intervention. The estimand is explicitly conditioning on evidence that $A_t=a_t$ from observational data, but counterfactually allowing the treatment to vary (to $a_t'$) to reason about the change in the mean. Another example, in the case of soft interventions, is when one may be interested in understanding the expected outcome using policy $\pi_e$ having observed outcomes using policy $\pi_b$ retrospectively, i.e. for the same set of patients treated with $\pi_b$ denoted as: $\mathbb{E}[Y_T^{\pi_e}(\bU)|Y_T^{\pi_{b}}]$. We show the non-identifiability of ETT as an example here in the binary setting (note that this is a parametric assumption) as derived in~\citep{shpitser2012effects} for the case where $Y_t$ and $A_t$ are binary in Equation~\ref{eq:ett2}. Note that Equation~\ref{eq:ett1} can be written using law of total expectation. Then the target causal query is given by Equation~\ref{eq:ett2}. 

\begin{align}\label{eq:ett2}
    \begin{split}
        \mathbb{E}[Y_t^{A_t=a_t'}|A_t=a_t] = \frac{\mathbb{E}[Y_t^{A_t=a_t'}] - \mathbb{E}[Y_t^{A_t=a_t'}|A_t=a_t']p(a_t')}{p(a_t)}
    \end{split}
\end{align}

 Thus to estimate ETT interventional quantities $\mathbb{E}[Y_t^{A_t=a_t'}]$ and $\mathbb{E}[Y_t^{A_t=a_t'}|a_t']$ are required, at least without further assumptions. Clearly, this estimand is not a function of observed variables directly, showing that such estimation is non-identifiable from historical data collected based on the previous medication. 
 However if some experimental data is available, that allows us to anchor treatments to $a_t'$, e.g. a randomized controlled trial, both of these quantities can be estimated, providing us with a way to counterfactually reason about the effect of alternative treatments on those who have been treated differently i.e. with $a_t$. In practice, some additional assumptions like monotonicity of the effects for binary treatment on the outcome can enable identifiability entirely from observational data and have been outlined in~\citet[Chapter 9]{pearl2009causality}. Identifiability with a general notion of monotonicity for discrete SCMs is outlined in~\citet{oberst2019counterfactual}. These identifiability results again require  ignorability/no unobserved confounding, thereby highlighting how these are closely tied to structural assumptions on the data-generating mechanisms. Further note that these identifiability results are applicable assuming treatments are binary and/or discrete. For other general estimands, it may be possible to get off-policy (i.e. from observational data) estimates of retrospective OPE estimands for specific data-generating assumptions, like no unobserved confounding and parametric assumptions like linearity.
 
These examples highlight that many (at times strong) assumptions are necessary to reliably solve OPE tasks, particularly retrospective OPE tasks. Challenges of identifiability can be further complex if the corresponding estimand is  individual-level and require estimating a complete probability distribution as opposed to expectations. 
For the most general dynamic data-generating processes, identifiability of the appropriate counterfactual estimands without parametric assumptions i.e. entirely based on the graphical structure has been outlined in~\citet{shpitser2012counterfactuals}. 

Formally, we are now in a position to define Counterfactual OPE and Interventional OPE. We focus on the most general OPE estimand (i.e. the probability distribution over target counterfactual/interventional quantity of interest) in each case and suggest that sub-population and/or population analogues can be correspondingly outlined. Note that purely in terms of the causal estimand, and the corresponding causal query, it is possible to unify these definitions (and notations) as is often the case in causality literature, we argue that there is significant merit in making these distinctions transparent, particularly for OPE tasks to explicitly contextualize contributions and ground the applicability of inferences made of OPE estimates.

\begin{definition}{Counterfactual Off-Policy Evaluation:}\label{def:counter_ope}
Counterfactual OPE corresponds to estimating the outcome distribution of a policy $\pi_e$ based on a data set of trajectories $\cH_{T}^{\pi_b} \triangleq \{S_0, A_1, Y_1, \cdots, S_{T-1}, A_T, Y_T\}$ generated independently by policy(ies) $\pi_b$ on units \emph{treated} with $\pi_b$ corresponding to the causal estimand $P(G(\cH_{T}^{\pi_e}(\bU)) \vert \cH_{T}^{\pi_b})$, where $\bU$ selects the appropriate target unit(s) of the counterfactual intervention $\pi_e$ and $G(\cdot)$ is a function of counterfactual outcomes/trajectories.
\end{definition}
We now define Interventional OPE as follows:

\begin{definition}{Interventional Off-Policy Evaluation:}\label{def:inter_ope}
Interventional OPE corresponds to estimating the outcome distribution of a policy $\pi_e$ based on a data set of trajectories $\cH_{T}^{\pi_b} \triangleq \{S_0, A_1, Y_1, \cdots, S_{T-1}, A_T, Y_T\}$ generated independently by policy(ies) $\pi_b$ on any prospective units treated with $\pi_e$ corresponding to the causal estimand $P(G(\cH_{T}^{\pi_e}) \vert \cH_{T}^{\pi_b})$ where $G(\cdot)$ is a function of interventional/prospective trajectories $\cH_T^{\pi_e}$.
\end{definition}


More generally, the two examples and our definitions highlight gap in ML specific OPE literature suggesting OPE literature has largely focused on i) interventional analysis through off-policy evaluation, ii) focusing on estimates of the population, such as expectations, as opposed to individuals or rarely sub-populations and iii) assuming identifiability from observational data, thereby making tacit assumptions about the underlying data-generating procedure~\citep{tennenholtz2019off, namkoong2020off, singh2021learning}. 
Applications like medical domains easily motivate the need for generalization along the lines summarized in Figure~\ref{fig:summary} as we are usually interested in evaluating patient level utility of policies, not just prospectively, but also retrospectively. In the following we discuss implications for OPE based on its formalization as a causal estimand. In particular, we  propose that these distinctions be routinely highlighted, thereby clearly exposing the underlying assumptions and hence applicability of corresponding estimands~\citep{tennenholtz2019off,singh2021learning,namkoong2020off,oberst2019counterfactual}. 
We demonstrate the main utility of this exposition and the proposed generalizations specifically along these factors of OPE has practical implications. More specifically, these implications become clear by tying the associated estimation challenges to specific sources of uncertainties in the system. 

\section{Sources of Uncertainty in OPE.}
\label{sec:uncertainty}


Generalizing OPE as a causal estimand naturally highlights two key technical challenges for progress in OPE. These challenges are i) identifiability from available observational data, ii) estimation challenges. These challenges essentially manifest as induced uncertainty in the OPE estimation process. The causal characterization is useful because it further allows us to distinguish the nature of this uncertainty, and helps outline the role of humans in aiding and improving both the above aspects, thereby systematically mitigating the uncertainty in OPE estimation. Thus, these distinctions are not slight and have significant implications on our attempts to answer fundamental questions about OPE. We first discuss the different sources of uncertainty induced by the general characterization. 
\begin{figure}[htbp!]
    \centering
    \includegraphics{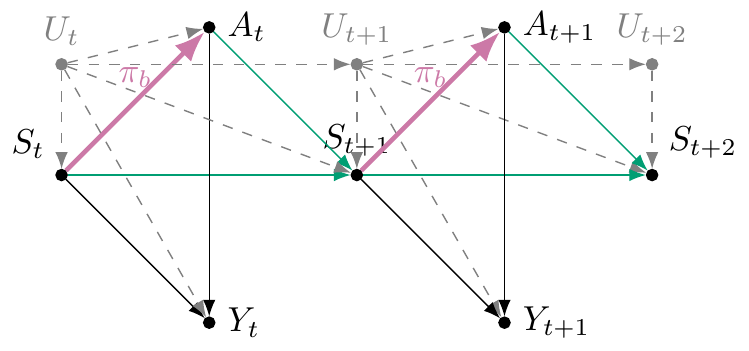}
    \caption{SCM analogue of a POMDP}
    \label{fig:pomdp}
\end{figure}
\begin{asparaenum}
{\bf \item {Identifiability.}} Knowledge of the data-generating process, usually provided by human expertise, including functional relationships can determine whether OPE using data collected from a behavior policy $\pi_b$ for another policy $\pi_e$ is possible. Challenges like {\emph{partial observability}} like in Figure~\ref{fig:pomdp} generally lead to non-identifiability for OPE, if we do not include additional assumptions. Non-identifiability implies the OPE estimate cannot be obtained even with infinite observational data from the behavior policy. This non-identifiability can be characterized as irreducible uncertainty in the OPE of interest. Characterizing this uncertainty received much less attention in ML literature for general causal inference~\cite{jesson2021quantifying}, let alone in OPE. Since Example 1 is a case of identifiable prospective OPE evaluation, there is no irreducible uncertainty induced due to non-identifiability. On the other hand, Example 2 suggests that without access to experimental data, the ETT estimate will have irreducible noise, and that experimentation should allow for  estimating $\mathbb{E}[Y^{a_t}]$ to specifically target this source of uncertainty. 

{\bf \item {Estimation.}} Once the appropriate observational and experimental data required for identifiable OPE are established, we may either have sufficient observational data, or collect some additional experimental data to enable estimation.  In practice, we often work with finite  samples, which are in-turn used to estimate the appropriate intermediate quantities that provide the OPE estimate. 
For example, Equation~\ref{eq:overlap}, requires reasonable estimates of the dynamics model, and the behavior policy. In Example 2, we will obtain finite sample estimates of $\mathbb{E}[Y^{A_t = a_t}]$ from experimental data. Finite sample issues induce additional source(s) of uncertainty, in some cases requiring large amounts of observational/experimental data for accurate estimation. This is true irrespective of the type of estimand, and might put additional demands on  data-collection needs. Since this uncertainty can be reduced by collecting additional data, this constitutes \emph{modeling} or \emph{epistemic} uncertainty.

Beyond structural assumptions about the data-generating process, parametric assumptions such as  stationarity of MDP dynamics further add to the source of modeling or epistemic uncertainty. 
For instance, \emph{non-stationary} exogenous variables often complicate model estimation, and can propagate uncertainty, thereby worsening how severe violations of assumptions like sufficient overlap~\citep{namkoong2020off,joshi2021sltd} are in the estimation process. These assumptions are a form of parametric assumptions that leave some sources of uncertainty irreducible (by absorbing non-stationarity into irreducible noise). Finally, lack of sufficient data, or inappropriately accounting for confounding variables (i.e. state representations) can also introduce biases in the estimate depending on the severity of violations of assumptions of ignorability and overlap~\citep{cinelli2020crash}. Issues of finite samples have only recently received attention and is necessary toward making causal inference a practical reality in ML and statistics~\citep{jung2021estimating,chernozhukov2018double}
\end{asparaenum}

Given these insights, we discuss dominant sources of uncertainty for different OPE estimands through our examples. 
\begin{asparaenum}
{\bf \item {Example 1:}} This is an identifiable (from observational data assumed to be already available) interventional/prospective OPE task. Thus, the key sources of uncertainty here are modeling uncertainty associated with estimating intermediate quantities like dynamics estimation, and modeling the outcome $p(Y_t|S_t, A_t)$ from finite samples.

{\bf \item {Example 2:}} This example of ETT requires both observational data to estimate the marginal $p(A_t)$, and  experimental data to estimate $\mathbb{E}[Y_t^{A_t=a_t'}]$ each inducing different sources of modeling uncertainty depending on the amount and quality of available data. If only observational data is available, then this OPE estimand is not identifiable, and the irreducible or aleatoric uncertainty corresponds to that induced due to our uncertainty of the knowledge of $\mathbb{E}[Y_t^{A_t=a_t'}]$.

\end{asparaenum}

We now discuss the implication of characterizing these sources of uncertainty. 
First is the need to obtain additional domain knowledge to reduce uncertainty associated with identifiability. In traditional causality literature, the general focus of establishing counterfactual identifiability has been on graphical criteria with less emphasis on parametric assumptions on the corresponding functional relationships between covariates of interest. 
Of course there are several examples where making parametric assumptions can aid identifiability, even if non-parametric identifiability does not hold~\cite{hernan2020causal, angrist1996identification,wright1921correlation}. 
Second is to  
resort to collecting experimental data, which in and of itself defeats the primary goal of OPE. 
More specifically, we discuss how humans can play a critical role for OPE where the causal estimand formalizes the type of human expertise and experimentation is required for practical OPE. We elaborate on these implications in detail in the following section. 

\section{The Role of Humans in OPE}
\label{sec:humans}
In this section, we illustrate how the causal estimand, and explicitly outlining the sources of uncertainty helps to precisely characterize how human feedback should be incorporated and what 
the role can fill in 
improving the reliability of OPE. 
In particular, 
the role of incorporating human expertise can be useful for: i) providing modeling assumptions (and relaxations) under which we can describe the right causal OPE estimand, 
ii) physical experimentation to aid identifiability, 
iii) mitigating sources of uncertainty to compensate for limited data, and 
iv) providing assumptions that enable to understand the conditions for the validity of the OPE estimand. In essence, most of these principled ways of incorporating human feedback can be chalked up to reducing different sources of uncertainty and improving the reliability of OPE estimates. We now elaborate on these distinct roles 
that human feedback can fulfill in OPE, while explicitly discussing the kind of benefit it may provide.

\paragraph{Parametric assumptions and modeling constraints.}\label{sec:learning} 
In many situations, making additional modeling assumptions can be sufficient and valid to establish some form of identifiability, depending on the application. Human experts can assess the applicability of such assumptions to the domain, as well as provide any additional constraints. For example, assumptions such as consistency, monotonicity~\citep[Chapter 9]{pearl2009causality}, or generalizations thereof are often used in epidemiological literature to estimate counterfactuals from observational data and have recently been explored in the context of OPE for model debugging~\citep{oberst2019counterfactual}. In {\bf{Example 2}}, assuming treatments are binary is a form of simple parametric assumption that can reduce the aleatoric uncertainty by reducing the number of interventional experiments that need to be conducted. These assumptions aid identifiability by making fundamental assumptions about the nature of counterfactuals, reducing aleatoric uncertainty. Additional knowledge like linearity and other functional assumptions, like the availability of noisy proxies of confounders~\citep{miao2018identifying,tennenholtz2019off}, or availability of instrumental variables can help constrain the amount of online experimentation required for OPE~\citep{xu2020learning}. These assumptions can help reduce modeling uncertainty, during estimation as well as reduce aleatoric uncertainty by constraining counterfactual distributions to specific functional forms. 

When populations are heterogeneous, human input can additionally provide such functional assumptions at a higher level of granularity by drawing from existing clinical knowledge. If the estimated dynamics in models are unreliable or may shift across domains, domain knowledge about the amount of shift can improve the robustness of OPE estimates in practice~\citep{singh2021learning}. In many cases, observational data may be insufficient, but experimentation is prohibitively costly or unethical. In such cases, domain experts can provide auxiliary model information like disease kinematics, or biophysical models collected from prior scientific experiments, as an alternative to simulating appropriate interventions \citep{adams2004dynamic,ribba2012tumor}. 
 This form of human or expert feedback can therefore reduce modeling uncertainty in the system. Most of these assumptions have been implicitly made to aid some form of identifiability in the most general sense and focused on population-level identifiability. 

\paragraph{Physical experimentation for aiding identifiability.} 
Establishing non-identifiability of a causal estimand corresponding to an OPE task can help identify the physical experimentation necessary to enable OPE for all cases we outline in Figure~\ref{fig:summary}. 
As we saw in {\bf{Example 2}},  
 OPE tasks requiring counterfactual reasoning, like ETT, might reveal the need for additional interventional/experimental data (or at the very least require domain expertise that further help establish identifiability). We demonstrated in Equation~\eqref{eq:ett2}, that in this case, a combination of i) binary treatment and outcome (parametric assumption a human can provide and justify) as well as ii) interventional/experimental data is necessary when no additional assumptions like modeling assumptions can be justified. Without experimentation, there is irreducible uncertainty in the system. 
Further, consider the POMDP in Figure~\ref{fig:pomdp} where the prospective/retrospective OPE is unidentifiable (in expectation or for individual-level OPE). To aid identifiability, one primarily needs to collect unobserved confounders. Thus, physical experimentation need not directly imply expensive experiments like a randomized controlled trial, but could also help if additional confounding variables could be collected to establish identifiability. Data collection for specific variables can be less prohibitive than an active collection of interventional quantities and has been explored recently for (conditional) average treatment effect estimation~\citep{wang2020confounding,bald2020causal}. Nonidentifiability exacerbated due to unobserved confounding and lack of the ability to experiment is the primary source of irreducible uncertainty or aleatoric uncertainty. Thus this form of human feedback can explicitly reduce aleatoric uncertainty. We argue that this is the main challenge where human input can further aid individual-level and retrospective OPE applications.

\paragraph{Mitigating estimation uncertainty from finite data.}
In many cases, while observational data collected from behavior policy might be sufficient for identifiability, the amount of data might be insufficient, significantly violating overlap assumptions which manifest in the increased variance of the OPE estimate. For instance, in Example 1, we may have very few samples collected from behavior policy, or the state representation may be high-dimensional. In these cases, humans may be able to analyze the validity of off-policy evaluation estimates \emph{posthoc} to, for instance, reduce our modeling uncertainty. To this end, \citet{pmlr-v119-gottesman20a} propose using humans to assess the validity of OPE estimates by manually analyzing 
the influential observations whose removal increases the estimation error. Similar approaches consider using human expertise to learn shaped control variates for variance reduction in OPE \citep{parbhooshaping2020} or using human expertise to define admissible rewards for high confidence OPE \citep{prasad2019defining}. This is another form of reducing stochasticity due to insufficient data i.e. reducing modeling uncertainty. While most existing work in this domain is applicable for prospective OPE tasks (and specifically {\bf{Example 1}}), focused on sub-population/population-level OPE, there is significant potential for human input to reduce modeling uncertainty for retrospective and individual-level OPE tasks.

\paragraph{Validating OPE estimates.}
Validation of OPE is a significant but critical challenge for practical utility. The potential for human feedback is the least well explored especially for retrospective OPE. 
Technically, sensitivity analysis models are the main modeling frameworks that can allow the incorporation of domain knowledge that characterizes the severity of unobserved confounding in the system. For instance, it can help validate prospective as well as retrospective OPE tasks for complex settings such as the POMDP (Figure~\ref{fig:pomdp}). Currently, the most commonly used sensitivity model known as the marginal sensitivity model, which constrains deviations to the propensity of treatment in the presence of unobserved confounders, and has been widely explored for population and sub-population-level prospective OPE in reinforcement learning~\citep{kallus2020confounding, namkoong2020off}. Model debugging of this kind has been explored for counterfactual OPE for discrete SCMs for managing sepsis among diabetic patients in~\citep{oberst2019counterfactual}.

\begin{figure}[t!]
    \centering
     \includegraphics[scale=0.5]{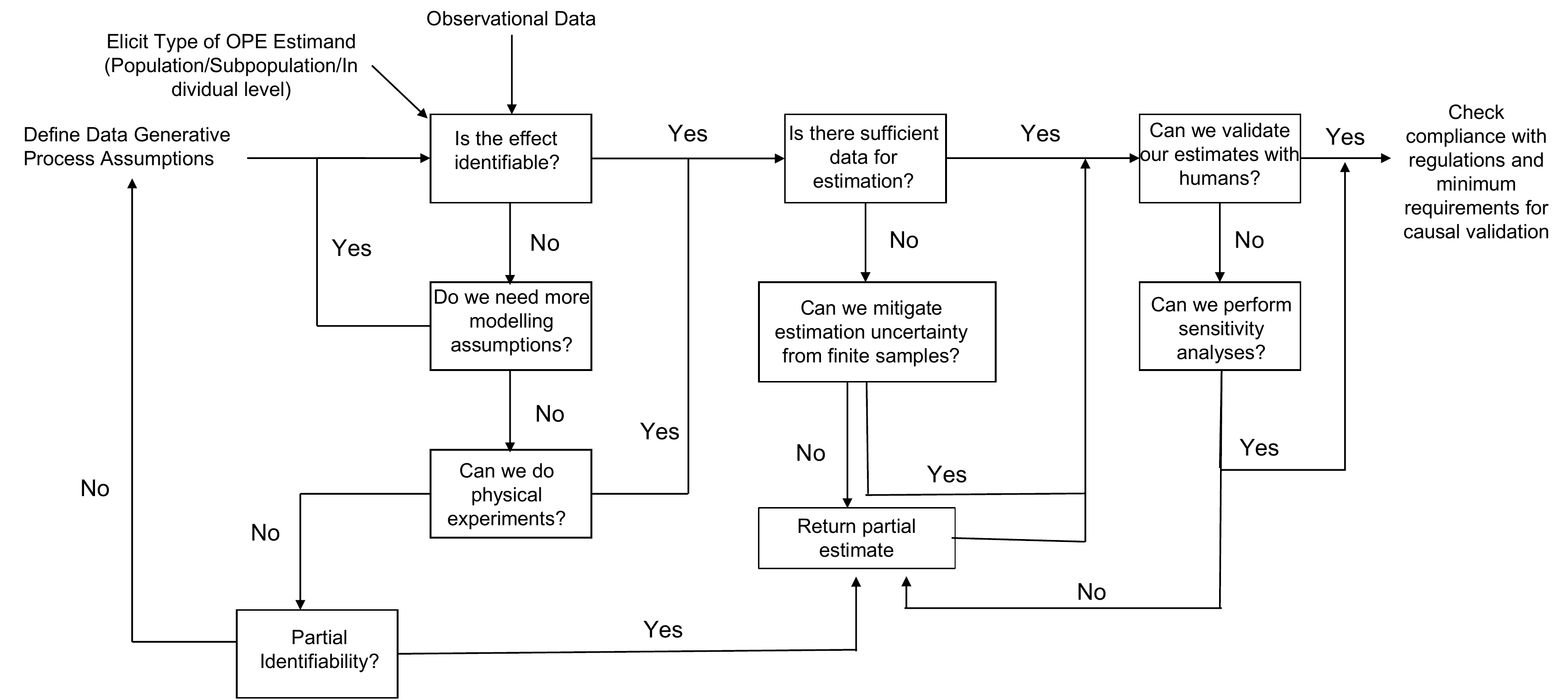}
    
    \caption{Roadmap for performing OPE in practice.}
    \label{fig:roadmap}
\end{figure} 

\section{A Roadmap for OPE in Practice}
\label{sec:discussion}
In this work, we motivate the need to generalize the purview of OPE in sequential decision-making, specifically the need to distinguish prospective versus retrospective OPE tasks conducted at the individual, sub-population, and population level. To operationalize the generalization, we argue that OPE tasks should be framed in the context of the data generating process and a corresponding causal estimand. We discuss how this operationalization can highlight and help mitigate fundamental OPE challenges related to identification, estimation, and principally outline the role of human input to make OPE tasks practical. In doing so we highlight many open questions. In this section, we conclude by providing a prescriptive roadmap for addressing OPE tasks in practice, and technical challenges to consider along the way. Our prescriptive roadmap is summarized in Figure~\ref{fig:roadmap}.

\paragraph{Elicit goals of OPE using a causal estimand.} By eliciting the appropriate causal estimand, we can transparently identify whether an OPE task is prospective/retrospective and required at the level of individuals, sub-populations, and population levels. More importantly, this allows us to consider whether existing observational data collected from a behavior policy and additional assumptions can be justified to estimate the corresponding estimand.  The corresponding assumptions in terms of the data-generating process (causal structure) and other  assumptions (such as parametric assumptions) help provide precise specifications of the claim and validity of the OPE estimand. 

\paragraph{Outline and justify nature of human input.}
The primary goal of OPE is to enable the evaluation of new policies based on observational data from behavior policies without additional experimentation. However, we highlight situations human input may be necessary and justified to solve specific OPE tasks. Here the form of human input can range from the need to conduct complex physical experimentation such as randomized controlled trials or in the form of data collection. This form of human input is explicitly aimed at mitigating non-identifiability issues. In some cases, human input might suffice to establish partial OPE estimates (bounds on the OPE estimate as opposed to point estimates). Even if an estimand is identifiable, we have to consider whether sufficient observational data is available to estimate the corresponding off-policy estimand. Here human input can help mitigate modeling uncertainty associated with estimation challenges by providing additional domain knowledge and would help improve the statistical properties of the corresponding estimators. 

\paragraph{Identify minimal validation requirements using causal desiderata.} The nature and type of validation required of OPE estimands are clear by focusing on the assumptions, the appropriate causal estimand, and the nature of human input incorporated to address identifiability and to obtain desirable estimation properties. In particular, these desiderata further identify the nature of validation, in the form of sensitivity analysis,  specifically of the type that invalidates assumptions justified for the domain, the amount and nature of unobserved confounding, or structural assumptions to clearly identify granular conditions under which the OPE estimand is meaningful. Finally, to reliably deploy policies that are evaluated in this manner, it is imperative to consider additional human factors that will allow for appropriate interpretation of recommendations inferred from the OPE estimates under the specifications outlined by our desiderata. 
 Note that most generally, assumptions of the data-generating process, as we have assumed throughout the draft, is a form of human expertise that allows to incorporate the most fundamental set of conditions of the domain under which the OPE estimand can be formulated. Using observational data as a means to learn such structures reliably is a nascent area called causal discovery or causal structure learning~\citep{glymour2019review}.

In conclusion, these considerations should move us towards a more rigorous way of performing OPE in practice and create a better understanding of those factors essential for making OPE reliable for practice. 

\bibliography{main}
\bibliographystyle{plainnat}

\end{document}

%% file: notation.tex
\def\to{{\,\rightarrow\,}}

\mathchardef\mhyphen="2D

\newcommand{\vertiii}[1]{{\left\vert\kern-0.25ex\left\vert\kern-0.25ex\left\vert #1
    \right\vert\kern-0.25ex\right\vert\kern-0.25ex\right\vert}}

\def\bv{{\mathbf{v}}}

\def\bF{{\mathbf{F}}}

\def\bP{{\mathbf{P}}}

\def\bS{{\mathbf{S}}}

\def\bU{{\mathbf{U}}}
\def\bV{{\mathbf{V}}}

\def\bX{{\mathbf{X}}}
\def\bY{{\mathbf{Y}}}

\def\cA{\mathcal{A}}

\def\cG{\mathcal{G}}
\def\cH{\mathcal{H}}

\def\cM{\mathcal{M}}
\def\cN{\mathcal{N}}

\def\cP{\mathcal{P}}

\def\cR{\mathcal{R}}
\def\cS{\mathcal{S}}





%% file: main.bbl
\begin{thebibliography}{45}
\providecommand{\natexlab}[1]{#1}
\providecommand{\url}[1]{\texttt{#1}}
\expandafter\ifx\csname urlstyle\endcsname\relax
  \providecommand{\doi}[1]{doi: #1}\else
  \providecommand{\doi}{doi: \begingroup \urlstyle{rm}\Url}\fi

\bibitem[Adams et~al.(2004)Adams, Banks, Kwon, and Tran]{adams2004dynamic}
Brian~M Adams, Harvey~T Banks, Hee-Dae Kwon, and Hien~T Tran.
\newblock Dynamic multidrug therapies for hiv: Optimal and sti control
  approaches.
\newblock \emph{Mathematical Biosciences \& Engineering}, 1\penalty0
  (2):\penalty0 223, 2004.

\bibitem[Angrist et~al.(1996)Angrist, Imbens, and
  Rubin]{angrist1996identification}
Joshua~D Angrist, Guido~W Imbens, and Donald~B Rubin.
\newblock Identification of causal effects using instrumental variables.
\newblock \emph{Journal of the American statistical Association}, 91\penalty0
  (434):\penalty0 444--455, 1996.

\bibitem[Bellman(1957)]{bellman1957markovian}
Richard Bellman.
\newblock A markovian decision process.
\newblock \emph{Journal of mathematics and mechanics}, 6\penalty0 (5):\penalty0
  679--684, 1957.

\bibitem[Chernozhukov et~al.(2018)Chernozhukov, Chetverikov, Demirer, Duflo,
  Hansen, Newey, and Robins]{chernozhukov2018double}
Victor Chernozhukov, Denis Chetverikov, Mert Demirer, Esther Duflo, Christian
  Hansen, Whitney Newey, and James Robins.
\newblock Double/debiased machine learning for treatment and structural
  parameters, 2018.

\bibitem[Cinelli et~al.(2020)Cinelli, Forney, and Pearl]{cinelli2020crash}
Carlos Cinelli, Andrew Forney, and Judea Pearl.
\newblock A crash course in good and bad controls.
\newblock \emph{Available at SSRN}, 3689437, 2020.

\bibitem[Dai et~al.(2020)Dai, Nachum, Chow, Li, Szepesv{\'a}ri, and
  Schuurmans]{dai2020coindice}
Bo~Dai, Ofir Nachum, Yinlam Chow, Lihong Li, Csaba Szepesv{\'a}ri, and Dale
  Schuurmans.
\newblock Coindice: Off-policy confidence interval estimation.
\newblock \emph{arXiv preprint arXiv:2010.11652}, 2020.

\bibitem[Glymour et~al.(2019)Glymour, Zhang, and Spirtes]{glymour2019review}
Clark Glymour, Kun Zhang, and Peter Spirtes.
\newblock Review of causal discovery methods based on graphical models.
\newblock \emph{Frontiers in genetics}, 10:\penalty0 524, 2019.

\bibitem[Gottesman et~al.(2019)Gottesman, Liu, Sussex, Brunskill, and
  Doshi-Velez]{gottesman2019combining}
Omer Gottesman, Yao Liu, Scott Sussex, Emma Brunskill, and Finale Doshi-Velez.
\newblock Combining parametric and nonparametric models for off-policy
  evaluation.
\newblock \emph{arXiv preprint arXiv:1905.05787}, 2019.

\bibitem[Gottesman et~al.(2020)Gottesman, Futoma, Liu, Parbhoo, Celi,
  Brunskill, and Doshi-Velez]{pmlr-v119-gottesman20a}
Omer Gottesman, Joseph Futoma, Yao Liu, Sonali Parbhoo, Leo Celi, Emma
  Brunskill, and Finale Doshi-Velez.
\newblock Interpretable off-policy evaluation in reinforcement learning by
  highlighting influential transitions.
\newblock In Hal~Daumé III and Aarti Singh, editors, \emph{Proceedings of the
  37th International Conference on Machine Learning}, volume 119 of
  \emph{Proceedings of Machine Learning Research}, pages 3658--3667. PMLR,
  13--18 Jul 2020.
\newblock URL \url{http://proceedings.mlr.press/v119/gottesman20a.html}.

\bibitem[Heckman(1992)]{heckman1992randomization}
James~J Heckman.
\newblock Randomization and social policy evaluation.
\newblock \emph{Evaluating welfare and training programs}, 1:\penalty0 201--30,
  1992.

\bibitem[Heckman and Vytlacil(2001)]{heckman2001policy}
James~J Heckman and Edward Vytlacil.
\newblock Policy-relevant treatment effects.
\newblock \emph{American Economic Review}, 91\penalty0 (2):\penalty0 107--111,
  2001.

\bibitem[Hern{\'a}n and Robins(2020)]{hernan2020causal}
Miguel~A Hern{\'a}n and James~M Robins.
\newblock Causal inference: what if, 2020.

\bibitem[Jesson et~al.(2021{\natexlab{a}})Jesson, Mindermann, Gal, and
  Shalit]{jesson2021quantifying}
Andrew Jesson, S{\"o}ren Mindermann, Yarin Gal, and Uri Shalit.
\newblock Quantifying ignorance in individual-level causal-effect estimates
  under hidden confounding.
\newblock \emph{arXiv preprint arXiv:2103.04850}, 2021{\natexlab{a}}.

\bibitem[Jesson et~al.(2021{\natexlab{b}})Jesson, Tigas, Amersfoort, Kirsch,
  Shalit, and Gal]{bald2020causal}
Andrew Jesson, Panagiotis Tigas, Joost~van Amersfoort, Andreas Kirsch, Uri
  Shalit, and Yarin Gal.
\newblock Causal-bald: Deep bayesian active learning of outcomes to infer
  treatment-effects from observational data, 2021{\natexlab{b}}.

\bibitem[Jiang and Li(2016)]{jiang2016doubly}
Nan Jiang and Lihong Li.
\newblock Doubly robust off-policy value evaluation for reinforcement learning.
\newblock In \emph{International Conference on Machine Learning}, pages
  652--661. PMLR, 2016.

\bibitem[Joshi* et~al.(2021)Joshi*, Parbhoo*, and Doshi-Velez]{joshi2021sltd}
Shalmali Joshi*, Sonali Parbhoo*, and Finale Doshi-Velez.
\newblock Interpretable learning-to-defer for sequential decision-making.
\newblock In \emph{ICML Workshop for Interpretable Machine Learning for
  Healthcare}, 2021.

\bibitem[Jung et~al.(2021)Jung, Tian, and Bareinboim]{jung2021estimating}
Yonghan Jung, Jin Tian, and Elias Bareinboim.
\newblock Estimating identifiable causal effects through double machine
  learning.
\newblock In \emph{Proceedings of the 35th AAAI Conference on Artificial
  Intelligence}, 2021.

\bibitem[Kallus and Zhou(2020)]{kallus2020confounding}
Nathan Kallus and Angela Zhou.
\newblock Confounding-robust policy evaluation in infinite-horizon
  reinforcement learning.
\newblock \emph{arXiv preprint arXiv:2002.04518}, 2020.

\bibitem[Levine et~al.(2020)Levine, Kumar, Tucker, and Fu]{levine2020offline}
Sergey Levine, Aviral Kumar, George Tucker, and Justin Fu.
\newblock Offline reinforcement learning: Tutorial, review, and perspectives on
  open problems.
\newblock \emph{arXiv preprint arXiv:2005.01643}, 2020.

\bibitem[Liao et~al.(2019)Liao, Klasnja, and Murphy]{liao2019off}
Peng Liao, Predrag Klasnja, and Susan Murphy.
\newblock Off-policy estimation of long-term average outcomes with applications
  to mobile health.
\newblock \emph{arXiv preprint arXiv:1912.13088}, 2019.

\bibitem[Mandel et~al.(2014)Mandel, Liu, Levine, Brunskill, and
  Popovic]{mandel2014offline}
Travis Mandel, Yun-En Liu, Sergey Levine, Emma Brunskill, and Zoran Popovic.
\newblock Offline policy evaluation across representations with applications to
  educational games.
\newblock In \emph{Proceedings of the 2014 international conference on
  Autonomous agents and multi-agent systems}, pages 1077--1084. International
  Foundation for Autonomous Agents and Multiagent Systems, 2014.

\bibitem[Miao et~al.(2018)Miao, Geng, and
  Tchetgen~Tchetgen]{miao2018identifying}
Wang Miao, Zhi Geng, and Eric~J Tchetgen~Tchetgen.
\newblock Identifying causal effects with proxy variables of an unmeasured
  confounder.
\newblock \emph{Biometrika}, 105\penalty0 (4):\penalty0 987--993, 2018.

\bibitem[Namkoong et~al.(2020)Namkoong, Keramati, Yadlowsky, and
  Brunskill]{namkoong2020off}
Hongseok Namkoong, Ramtin Keramati, Steve Yadlowsky, and Emma Brunskill.
\newblock Off-policy policy evaluation for sequential decisions under
  unobserved confounding.
\newblock \emph{arXiv preprint arXiv:2003.05623}, 2020.

\bibitem[Oberst and Sontag(2019)]{oberst2019counterfactual}
Michael Oberst and David Sontag.
\newblock Counterfactual off-policy evaluation with gumbel-max structural
  causal models.
\newblock In \emph{International Conference on Machine Learning}, pages
  4881--4890. PMLR, 2019.

\bibitem[Parbhoo et~al.(2017)Parbhoo, Bogojeska, Zazzi, Roth, and
  Doshi-Velez]{parbhoo2017combining}
Sonali Parbhoo, Jasmina Bogojeska, Maurizio Zazzi, Volker Roth, and Finale
  Doshi-Velez.
\newblock Combining kernel and model based learning for hiv therapy selection.
\newblock \emph{AMIA Summits on Translational Science Proceedings},
  2017:\penalty0 239, 2017.

\bibitem[Parbhoo et~al.(2018)Parbhoo, Gottesman, Ross, Komorowski, Faisal, Bon,
  Roth, and Doshi-Velez]{parbhoo2018improving}
Sonali Parbhoo, Omer Gottesman, Andrew~Slavin Ross, Matthieu Komorowski, Aldo
  Faisal, Isabella Bon, Volker Roth, and Finale Doshi-Velez.
\newblock Improving counterfactual reasoning with kernelised dynamic mixing
  models.
\newblock \emph{PloS one}, 13\penalty0 (11):\penalty0 e0205839, 2018.

\bibitem[Parbhoo et~al.(2020)Parbhoo, Gottesman, and
  Doshi-Velez]{parbhooshaping2020}
Sonali Parbhoo, Omer Gottesman, and Finale Doshi-Velez.
\newblock Shaping control variates for off-policy evaluation.
\newblock In \emph{Offline Reinforcement Learning Workshop at Neural
  Information Processing Systems (NeurIPS)}, 2020.

\bibitem[Pearl(2009)]{pearl2009causality}
Judea Pearl.
\newblock \emph{Causality}.
\newblock Cambridge university press, 2009.

\bibitem[Prasad et~al.(2019)Prasad, Engelhardt, and
  Doshi-Velez]{prasad2019defining}
Niranjani Prasad, Barbara~E Engelhardt, and Finale Doshi-Velez.
\newblock Defining admissible rewards for high confidence policy evaluation.
\newblock \emph{arXiv preprint arXiv:1905.13167}, 2019.

\bibitem[Precup(2000{\natexlab{a}})]{precup2000}
Doina Precup.
\newblock Eligibility traces for off-policy policy evaluation.
\newblock \emph{Computer Science Department Faculty Publication Series},
  page~80, 2000{\natexlab{a}}.

\bibitem[Precup(2000{\natexlab{b}})]{precup2000eligibility}
Doina Precup.
\newblock Eligibility traces for off-policy policy evaluation.
\newblock \emph{Computer Science Department Faculty Publication Series},
  page~80, 2000{\natexlab{b}}.

\bibitem[Ribba et~al.(2012)Ribba, Kaloshi, Peyre, Ricard, Calvez, Tod,
  {\v{C}}ajavec-Bernard, Idbaih, Psimaras, Dainese, et~al.]{ribba2012tumor}
Benjamin Ribba, Gentian Kaloshi, Mathieu Peyre, Damien Ricard, Vincent Calvez,
  Michel Tod, Branka {\v{C}}ajavec-Bernard, Ahmed Idbaih, Dimitri Psimaras,
  Linda Dainese, et~al.
\newblock A tumor growth inhibition model for low-grade glioma treated with
  chemotherapy or radiotherapy.
\newblock \emph{Clinical Cancer Research}, 2012.

\bibitem[Richardson and Robins(2013)]{richardson2013single}
Thomas~S Richardson and James~M Robins.
\newblock Single world intervention graphs (swigs): A unification of the
  counterfactual and graphical approaches to causality.
\newblock \emph{Center for the Statistics and the Social Sciences, University
  of Washington Series. Working Paper}, 128\penalty0 (30):\penalty0 2013, 2013.

\bibitem[Richens et~al.(2019)Richens, Lee, and
  Johri]{richens2019counterfactual}
Jonathan~G Richens, Ciar{\'a}n~M Lee, and Saurabh Johri.
\newblock Counterfactual diagnosis.
\newblock \emph{arXiv preprint arXiv:1910.06772}, 2019.

\bibitem[Shpitser and Pearl(2012{\natexlab{a}})]{shpitser2012counterfactuals}
Ilya Shpitser and Judea Pearl.
\newblock What counterfactuals can be tested.
\newblock \emph{arXiv preprint arXiv:1206.5294}, 2012{\natexlab{a}}.

\bibitem[Shpitser and Pearl(2012{\natexlab{b}})]{shpitser2012effects}
Ilya Shpitser and Judea Pearl.
\newblock Effects of treatment on the treated: Identification and
  generalization.
\newblock \emph{arXiv preprint arXiv:1205.2615}, 2012{\natexlab{b}}.

\bibitem[Silver et~al.(2013)Silver, Newnham, Barker, Weller, and
  McFall]{silver2013concurrent}
David Silver, Leonard Newnham, David Barker, Suzanne Weller, and Jason McFall.
\newblock Concurrent reinforcement learning from customer interactions.
\newblock In \emph{International Conference on Machine Learning}, pages
  924--932, 2013.

\bibitem[Singh et~al.(2021)Singh, Joshi, Doshi-Velez, and
  Lakkaraju]{singh2021learning}
Harvineet Singh, Shalmali Joshi, Finale Doshi-Velez, and Himabindu Lakkaraju.
\newblock Learning under adversarial and interventional shifts.
\newblock \emph{arXiv preprint arXiv:2103.15933}, 2021.

\bibitem[Sutton and Barto(2018)]{sutton2018reinforcement}
R.S. Sutton and A.G. Barto.
\newblock \emph{Reinforcement Learning, second edition: An Introduction}.
\newblock Adaptive Computation and Machine Learning series. MIT Press, 2018.
\newblock ISBN 9780262352703.
\newblock URL \url{https://books.google.com/books?id=uWV0DwAAQBAJ}.

\bibitem[Tennenholtz et~al.(2019)Tennenholtz, Mannor, and
  Shalit]{tennenholtz2019off}
Guy Tennenholtz, Shie Mannor, and Uri Shalit.
\newblock Off-policy evaluation in partially observable environments.
\newblock \emph{arXiv preprint arXiv:1909.03739}, 2019.

\bibitem[Thomas and Brunskill(2016)]{thomas2016data}
Philip Thomas and Emma Brunskill.
\newblock Data-efficient off-policy policy evaluation for reinforcement
  learning.
\newblock In \emph{International Conference on Machine Learning}, pages
  2139--2148. PMLR, 2016.

\bibitem[Valeri et~al.(2016)Valeri, Chen, Garcia-Albeniz, Krieger, VanderWeele,
  and Coull]{valeri2016role}
Linda Valeri, Jarvis~T Chen, Xabier Garcia-Albeniz, Nancy Krieger, Tyler~J
  VanderWeele, and Brent~A Coull.
\newblock The role of stage at diagnosis in colorectal cancer black--white
  survival disparities: a counterfactual causal inference approach.
\newblock \emph{Cancer Epidemiology and Prevention Biomarkers}, 25\penalty0
  (1):\penalty0 83--89, 2016.

\bibitem[Wang et~al.(2020)Wang, Yi, Joshi, and Ghassemi]{wang2020confounding}
Shirly Wang, Seung~Eun Yi, Shalmali Joshi, and Marzyeh Ghassemi.
\newblock Confounding feature acquisition for causal effect estimation.
\newblock In \emph{Machine Learning for Health}, pages 379--396. PMLR, 2020.

\bibitem[Wright(1921)]{wright1921correlation}
Sewall Wright.
\newblock Correlation and causation.
\newblock \emph{Journal of Agricultural Research}, 1921.

\bibitem[Xu et~al.(2020)Xu, Chen, Srinivasan, de~Freitas, Doucet, and
  Gretton]{xu2020learning}
Liyuan Xu, Yutian Chen, Siddarth Srinivasan, Nando de~Freitas, Arnaud Doucet,
  and Arthur Gretton.
\newblock Learning deep features in instrumental variable regression.
\newblock \emph{arXiv preprint arXiv:2010.07154}, 2020.

\end{thebibliography}
